\documentclass[11pt]{article}

\usepackage[utf8]{inputenc}
\usepackage[T1]{fontenc}
\usepackage{lmodern}
\usepackage[a4paper,margin=1in]{geometry}
\usepackage{booktabs}
\usepackage{amsmath}
\usepackage{array}
\usepackage{enumitem}
\usepackage{hyperref}
\usepackage[round]{natbib}

\hypersetup{
    colorlinks=true,
    linkcolor=blue,
    citecolor=blue,
    urlcolor=blue
}

\title{Building and Evaluating a Synthetic Bengali Speech Resource for Telecom Customer Care}

\author{
Kawshik Kumar Paul$^{1}$ \quad Md. Nafiul Alam Fuji$^{2}$\\
{\small Department of Computer Science and Engineering}\\
{\small Bangladesh University of Engineering and Technology (BUET)}\\
{\small $^{1}$\texttt{kawshikbuet17@gmail.com} \quad $^{2}$\texttt{nafiul.fuji@gmail.com}}
}

\date{}

\begin{document}

\maketitle

\begin{abstract}
Speech systems used in customer-facing applications often require domain-specific language coverage. We present a synthetic Bengali speech dataset for telecom customer-care scenarios. The dataset contains 10,000 audio-text pairs, approximately 26.82 hours of 24 kHz speech, and predefined train, validation, and test splits of 9,000, 500, and 500 examples. It is publicly released on Hugging Face under the CC-BY-4.0 license. The speech was generated with OmniVoice in voice-cloning mode using a real female reference recording and transcript, with bfloat16 precision, 16 diffusion sampling steps, and a speaking-rate control value of 1.0. Along with the original Bengali text, the dataset provides a normalized transcript field designed for ASR/STT training and evaluation. We report an automatic intelligibility check over all 10,000 samples using a domain-adapted Whisper ASR model fine-tuned from \path{bengaliAI/tugstugi_bengaliai-regional-asr_whisper-medium}, along with a manual listening check on selected samples. The evaluation gives an average WER of 2.54\%, an average CER of 0.59\%, and median WER and CER values of 0.00\%. These results suggest strong text-audio consistency under the selected automatic evaluation pipeline, while the paper also discusses the limitations of synthetic speech and STT-based evaluation.
\end{abstract}

\section{Introduction}

Many practical speech systems operate in specific domains where the vocabulary, phrasing, and user intents are repeated. Telecom and customer-care interactions are good examples. A user may ask about a failed recharge, an OTP delay, a package activation issue, a balance mismatch, SIM or account blocking, refund status, registration problems, or mobile banking verification. These situations require speech systems that can handle domain-specific expressions in a reliable way.

This paper presents \textit{Bengali Telecom Customer Care Synthetic Speech Dataset}, a synthetic Bengali speech resource that we released on Hugging Face \citep{kawshik_kumar_paul_2026}. The dataset contains 10,000 synthetic speech samples generated from telecom and customer-care style text prompts. It includes both the original text used for speech generation and a normalized text field intended for ASR/STT training and evaluation.

The dataset was generated using OmniVoice \citep{zhu2026omnivoice}. We use OmniVoice as a speech generation system; the contribution of this work is not a new TTS architecture. The contribution is the construction, release, documentation, and evaluation of a Bengali domain-specific synthetic speech resource.

The main contributions of this paper are:

\begin{itemize}[leftmargin=*]
    \item We release a 10,000-sample synthetic Bengali speech dataset focused on telecom and customer-care style utterances.
    \item We provide both original and normalized Bengali text fields to support TTS and ASR/STT workflows.
    \item We document the generation setup, including OmniVoice voice-cloning mode, bfloat16 precision, 16 diffusion sampling steps, and 24 kHz output audio.
    \item We evaluate all 10,000 samples using a fine-tuned Whisper-based STT intelligibility check and report WER/CER summary results.
    \item We discuss practical limitations and ethical considerations for using synthetic voice-cloned speech.
\end{itemize}

\section{Related Work}

\subsection{Open speech resources}

Open speech corpora have played an important role in multilingual speech research. Common Voice is a massively multilingual speech corpus created through crowdsourced speech collection and validation, and it is designed primarily for automatic speech recognition research and development \citep{ardila2020commonvoice}. Resources of this type show the importance of public speech datasets for building and evaluating speech systems.

The dataset in this paper differs from crowdsourced human-recorded corpora in two major ways. First, it is synthetic rather than human-recorded. Second, it is domain-specific rather than broad-domain. Its focus is Bengali telecom and customer-care style speech.

\subsection{Synthetic speech as a dataset resource}

Synthetic speech has also been used as a resource-building strategy. For example, the CVSS corpus uses TTS to construct multilingual speech-to-speech translation data from translated text \citep{jia2022cvss}. This shows that synthetic speech can be useful when the generation method and limitations are clearly stated.

The present work follows the same general direction of using synthetic speech as a research resource, but the target is different. Instead of speech translation, this dataset targets Bengali telecom/customer-care speech and provides paired audio, original text, and normalized transcripts for speech experiments.

\subsection{OmniVoice}

OmniVoice is a massively multilingual zero-shot TTS model that scales to more than 600 languages and directly maps text to multi-codebook acoustic tokens using a diffusion language model-style discrete non-autoregressive architecture \citep{zhu2026omnivoice}. The model supports voice cloning, which makes it suitable for generating synthetic speech from a reference recording. In this work, OmniVoice is used as the generation backend for the dataset.

\subsection{Whisper-based ASR}

Whisper is an automatic speech recognition system trained with large-scale weak supervision and evaluated across multilingual and multitask settings \citep{radford2023whisper}. For the automatic evaluation in this paper, we used a domain-adapted Whisper ASR model fine-tuned for our Bengali telecom/customer-care evaluation setting. The model was initialized from \path{bengaliAI/tugstugi_bengaliai-regional-asr_whisper-medium}, a Whisper-based Bengali ASR model released on Hugging Face. The model card describes it as being trained on regional Bengali speech data covering 10 dialects and refers to the BEN10 dataset \citep{tugstugi_bengaliai-regional-asr_whisper-medium}.

We further fine-tuned this model using additional open-source Bengali speech data and privately recorded Bengali speech data containing domain-specific telecom/customer-care expressions. The fine-tuning data was filtered using a separate custom wav2vec2-based ASR system, retaining samples with WER $\leq 10\%$. We refer to this evaluator as our fine-tuned Tugstugi Whisper model in the rest of the paper.

The evaluator is used only as an automatic proxy for text-audio consistency. Since it was adapted to the target language and domain, the reported WER/CER values should not be interpreted as a fully independent benchmark of the dataset.

\subsection{WER and CER for ASR evaluation}

Word error rate (WER) is a common metric for ASR evaluation. It counts substitutions, insertions, and deletions at the word level \citep{huggingface_audio_eval}. Character error rate (CER) applies the same idea at the character level \citep{huggingface_audio_eval}. CER can be useful in multilingual ASR evaluation because word-level tokenization and writing-system differences can affect WER interpretation \citep{dk2024cer}. For this reason, we report both WER and CER in our automatic evaluation.

\section{Dataset Overview}

\subsection{Dataset content}

The dataset contains synthetic Bengali speech generated from telecom and customer-care style prompts. The dataset card lists Bengali ASR/STT, Bengali TTS, speech-to-text preprocessing, telecom/customer-care domain adaptation, and synthetic speech research as intended uses \citep{kawshik_kumar_paul_2026}.

Table~\ref{tab:dataset_overview} summarizes the released dataset.

\begin{table}[!htbp]
\centering
\begin{tabular}{ll}
\toprule
Property & Value \\
\midrule
Dataset name & Bengali Telecom Customer Care Synthetic Speech Dataset \\
Language & Bengali \\
Language code & bn \\
Domain & Telecom and customer-care style utterances \\
Speech type & Synthetic \\
Total examples & 10,000 \\
Approximate total duration & 26.82 hours \\
Sample rate & 24,000 Hz \\
License & CC-BY-4.0 \\
Repository & Hugging Face dataset repository \\
\bottomrule
\end{tabular}
\caption{Overview of the released dataset \citep{kawshik_kumar_paul_2026}.}
\label{tab:dataset_overview}
\end{table}

\subsection{Dataset splits}

The dataset is distributed with train, validation, and test splits. The split sizes are shown in Table~\ref{tab:splits}.

\begin{table}[!htbp]
\centering
\begin{tabular}{lr}
\toprule
Split & Number of examples \\
\midrule
Train & 9,000 \\
Validation & 500 \\
Test & 500 \\
\midrule
Total & 10,000 \\
\bottomrule
\end{tabular}
\caption{Dataset split sizes \citep{kawshik_kumar_paul_2026}.}
\label{tab:splits}
\end{table}

\subsection{Metadata fields}

Each metadata row contains the fields shown in Table~\ref{tab:metadata_fields}. The metadata is intentionally kept simple so that users can load it directly for TTS, ASR/STT, or preprocessing experiments.

\begin{table}[!htbp]
\centering
\begin{tabular}{ll}
\toprule
Field & Description \\
\midrule
\texttt{file\_name} & Relative path to the audio file \\
\texttt{text} & Original Bengali text used for TTS generation \\
\texttt{text\_normalized} & Normalized Bengali text suitable for ASR/STT training \\
\texttt{language} & Language code, \texttt{bn} for Bengali \\
\texttt{is\_synthetic} & Whether the sample is synthetic \\
\texttt{sample\_rate} & Audio sample rate \\
\texttt{duration\_sec} & Audio duration in seconds \\
\bottomrule
\end{tabular}
\caption{Metadata fields in the released dataset \citep{kawshik_kumar_paul_2026}.}
\label{tab:metadata_fields}
\end{table}

Per-sample generation runtime, STT transcript, WER, CER, and inference-time logs are not included in the main dataset metadata because these values are evaluation artifacts rather than dataset labels. The released metadata keeps only the fields needed for standard TTS and ASR/STT use, while aggregate evaluation results are reported separately in Section~\ref{sec:evaluation}.

\section{Text Preparation and Normalization}

The dataset provides two text fields: \texttt{text} and \texttt{text\_normalized}. The \texttt{text} field stores the original Bengali sentence used during speech generation. The \texttt{text\_normalized} field stores a cleaner transcript form intended for ASR/STT training and evaluation.

This separation is useful because TTS and ASR do not always need exactly the same text representation. In TTS, the input may be a readable written sentence. In ASR/STT, the target transcript should avoid formatting differences that do not correspond to clearly audible differences. For example, punctuation, spacing, abbreviation formatting, and some Bengali spelling variants may affect text matching even when the audio is understandable. Normalization therefore helps reduce metric inflation caused by written-form differences rather than speech errors.

The dataset card recommends using \texttt{text} as the TTS input and \texttt{text\_normalized} as the ASR/STT target transcript \citep{kawshik_kumar_paul_2026}. We follow the same convention in the evaluation section of this paper.

\subsection{Normalization procedure}

The normalized transcript field was prepared using rule-based text cleanup. The normalization mainly included punctuation cleanup, collapsing repeated punctuation marks, standardizing similar punctuation variants, removing unnecessary spacing, normalizing abbreviation formatting, handling Bengali number-word variants, and standardizing common written-form variants in borrowed telecom terms. The purpose of this normalization was to reduce ASR/STT evaluation mismatches caused by written-form differences rather than audible speech differences. No semantic changes were intentionally introduced during normalization.

\section{Synthetic Speech Generation}

All speech in the dataset is synthetic. The dataset card states that the audio was generated using the OmniVoice TTS system in voice-cloning mode, and that a real female voice recording with its transcript was used as the reference voice for cloning \citep{kawshik_kumar_paul_2026}. It also states that the dataset does not contain real customer-care recordings, real customer conversations, or real user audio \citep{kawshik_kumar_paul_2026}.

The documented generation settings are shown in Table~\ref{tab:generation_config}.

\begin{table}[!htbp]
\centering
\begin{tabular}{ll}
\toprule
Parameter & Value \\
\midrule
TTS system & OmniVoice \\
Generation mode & Voice cloning \\
Reference source & Real female voice recording and transcript \\
Precision & bfloat16 \\
Diffusion sampling steps & 16 \\
Speaking-rate control & speed = 1.0 \\
Output sample rate & 24,000 Hz \\
\bottomrule
\end{tabular}
\caption{Speech generation configuration documented in the dataset card \citep{kawshik_kumar_paul_2026}.}
\label{tab:generation_config}
\end{table}

\section{STT-based Intelligibility Evaluation}
\label{sec:evaluation}

\subsection{Evaluation design}

To estimate how well the generated speech matches the intended Bengali text, we evaluated all 10,000 generated samples using our fine-tuned Tugstugi Whisper model. For each sample, the generated audio was transcribed by the ASR evaluator, and the resulting transcript was compared with the corresponding \texttt{text\_normalized} field.

The evaluation covers every sample in the dataset. The scores are used as automatic proxy indicators of intelligibility and text-audio consistency. They should not be interpreted as human naturalness ratings or as direct labels of sample quality.

Because the ASR evaluator was adapted for the Bengali telecom/customer-care setting, the reported WER/CER values should be interpreted as an automatic consistency check under this evaluation pipeline rather than as a fully independent benchmark. The exact fine-tuning data, decoding configuration, and complete normalization script would be needed as a separate evaluation artifact for full reproducibility. In this paper, we report the dataset-level summary statistics from the completed evaluation.

\subsection{Manual listening check}

In addition to the ASR-based consistency check, we performed an informal manual listening check on a subset of generated samples. Human reviewers listened to selected audio files and compared them with the intended text to identify obvious text-audio mismatches, pronunciation problems, incomplete speech, or unnatural repetitions. This check was used as a qualitative sanity check rather than as a formal mean-opinion-score evaluation.

Because the manual check was not designed as a controlled listening study, we do not report MOS or listener-preference scores. The automatic WER/CER evaluation remains the main dataset-level consistency measure, while the manual check provides additional qualitative evidence that selected generated samples were understandable and aligned with the intended Bengali text.

\subsection{Metrics}

For both WER and CER, errors are computed using substitutions, insertions, and deletions. If $S$ is the number of substitutions, $I$ is the number of insertions, $D$ is the number of deletions, and $N$ is the number of units in the reference sequence, then the error rate is:

\begin{equation}
\mathrm{ErrorRate} = \frac{S + I + D}{N}.
\end{equation}

For WER, the units are words. For CER, the units are characters. A score of 0 indicates an exact match between the reference and the prediction. Lower values indicate better agreement.

\subsection{Results}

Table~\ref{tab:stt_results} shows the STT-based evaluation results. The average WER is 2.54\%, and the average CER is 0.59\%. The median WER and CER are both 0.00\%.

\begin{table}[!htbp]
\centering
\begin{tabular}{lr}
\toprule
Metric & Value \\
\midrule
Total samples & 10,000 \\
Scored samples & 10,000 \\
Average WER & 2.54\% \\
Median WER & 0.00\% \\
Average CER & 0.59\% \\
Median CER & 0.00\% \\
\bottomrule
\end{tabular}
\caption{STT-based evaluation summary. WER and CER were computed by comparing transcripts from our fine-tuned Tugstugi Whisper model against \texttt{text\_normalized}.}
\label{tab:stt_results}
\end{table}

The zero median values indicate that at least half of the generated samples were transcribed exactly after normalization. The non-zero average values show that a smaller subset of utterances contained mismatches under the automatic STT-based evaluation pipeline.

\subsection{Discussion}

The low average CER suggests strong character-level agreement between the generated speech and the normalized reference text. The WER is higher than CER, which is expected because a single word-level mismatch can increase WER more strongly than CER. For Bengali and other languages where spelling, spacing, morphology, and tokenization can influence word boundaries, CER provides a useful complementary view to WER \citep{dk2024cer}.

Manual inspection of selected non-zero WER/CER cases suggested that some mismatches were due to minor written-form differences rather than clear speech-generation errors. Examples include Bengali number-word variants, abbreviation spacing, and orthographic alternatives. Therefore, the reported WER/CER values should be interpreted as automatic proxy indicators of text-audio consistency, not as direct labels of sample quality.

These results should be read as evidence of text-audio consistency under an automatic evaluation pipeline. They do not prove that the synthetic speech is natural, human-like, or suitable for all deployment conditions. Human listening tests would be needed to measure perceived naturalness, speaker similarity, and listener preference.

\section{Use Cases}

The dataset is designed for several practical research and development use cases.

\paragraph{ASR and STT.}
Users can train or evaluate Bengali ASR/STT systems using the mapping \texttt{audio} $\rightarrow$ \texttt{text\_normalized}. The normalized transcript field is recommended because it is designed to reduce formatting mismatch in the target text.

\paragraph{Text-to-speech.}
Users can use the mapping \texttt{text} $\rightarrow$ \texttt{audio} for TTS experiments. Since the audio is synthetic, the dataset may be useful for controlled experiments, bootstrapping, or domain adaptation.

\paragraph{Telecom and customer-care domain adaptation.}
The utterances follow customer-care style patterns, making the dataset suitable for experiments where broad-domain speech data may not contain enough telecom/customer-care phrasing.

\paragraph{Synthetic speech research.}
Because the dataset is explicitly synthetic and includes a generation disclosure, it can be used for experiments on synthetic speech utility, ASR robustness to generated speech, and synthetic-to-real transfer.

\section{Limitations}

The dataset has several limitations.

First, it is synthetic. The dataset card states that it does not contain real customer-care recordings, real customer conversations, or real user audio \citep{kawshik_kumar_paul_2026}. Models trained only on this dataset may not fully generalize to real-world speech, noisy phone calls, spontaneous conversation, different microphones, real accents, or background noise. The dataset card also recommends combining this resource with real human-recorded Bengali speech for production-quality ASR or TTS \citep{kawshik_kumar_paul_2026}.

Second, the dataset uses voice cloning from a real female reference recording. This limits speaker diversity compared with a multi-speaker corpus. Future versions can improve this by adding multiple reference voices or voice-design variants, provided that consent and licensing are handled clearly.

Third, the STT-based evaluation is automatic. WER and CER can help estimate text-audio consistency, but they do not measure all aspects of audio quality. A sample can have low WER/CER and still sound unnatural. Conversely, a sample can be understandable to a human listener but receive a higher STT error because of ASR limitations.

Fourth, normalization choices affect WER and CER. If a different normalization pipeline is used, the metric values may change. For this reason, the evaluation script and normalization rules should be treated as part of the reproducibility package.

\section{Ethical Considerations}

The dataset avoids the direct privacy risks of releasing real customer-service calls because all audio samples are synthetic. The dataset card explicitly states that it does not contain real customer-care recordings, real customer conversations, or real user audio \citep{kawshik_kumar_paul_2026}.

At the same time, voice cloning requires careful handling. The dataset was generated using a real female voice recording and its transcript as a reference for cloning \citep{kawshik_kumar_paul_2026}. The reference recording used for voice cloning was used with permission from the speaker. Any use of cloned voice data should consider consent, speaker rights, impersonation risks, and possible misuse. This dataset should not be used for deception, speaker impersonation, fraud, or misrepresentation of a real person.

Users should also be careful when applying synthetic speech data to production systems. Synthetic data can be useful for research and development, but production systems should be evaluated with realistic speech data and appropriate user-safety checks.

\section{Conclusion}

We presented a synthetic Bengali speech resource for telecom customer-care scenarios. The dataset contains 10,000 synthetic audio-text pairs, approximately 26.82 hours of speech, and 24 kHz audio. It includes both original Bengali text and normalized transcripts, making it usable for both TTS and ASR/STT experiments.

The dataset was generated using OmniVoice in voice-cloning mode and released publicly on Hugging Face under the CC-BY-4.0 license. We evaluated all 10,000 samples using a fine-tuned Tugstugi Whisper-based intelligibility check. The evaluation produced an average WER of 2.54\%, an average CER of 0.59\%, and median WER and CER values of 0.00\%.

These results suggest that the dataset has strong text-audio consistency under the selected automatic evaluation pipeline. Future work includes releasing detailed evaluation artifacts, adding speaker diversity, comparing voice-cloning and voice-design generation, conducting human listening tests, and combining the dataset with real Bengali speech for more robust speech system development.

\section*{Data Availability}

The dataset is publicly available at: 
\url{https://huggingface.co/datasets/kawshikbuet17/bengali-telecom-customer-care-speech}

\bibliographystyle{plainnat}
\bibliography{references}

\end{document}